\documentclass[conference]{IEEEtran}
\usepackage{times}
\usepackage{booktabs}
\usepackage[numbers]{natbib}
\usepackage{multicol}
\usepackage[bookmarks=true]{hyperref}
\usepackage{amsmath} 
\usepackage{xcolor}
\usepackage{graphicx}
\usepackage{caption}
\usepackage{subcaption}
\usepackage{bm}
\usepackage{bbm}
\newcommand{\boxminus}{\ominus}

\begin{document}

\title{Generalizing from References using a Multi-Task Reference and Goal-Driven RL Framework}


\author{\authorblockN{Jiashun Wang\textsuperscript{1,2} , M. Eva Mungai\textsuperscript{1} , He Li\textsuperscript{1}, Jean Pierre Sleiman\textsuperscript{1}, Jessica Hodgins\textsuperscript{1,2}, Farbod Farshidian\textsuperscript{1}} 
\authorblockA{\textsuperscript{1}RAI Institute \quad \textsuperscript{2}Carnegie Mellon University}
}



%

\makeatletter
\let\@oldmaketitle\@maketitle
    \renewcommand{\@maketitle}{\@oldmaketitle
    \centering
    \includegraphics[width=1.0\textwidth]{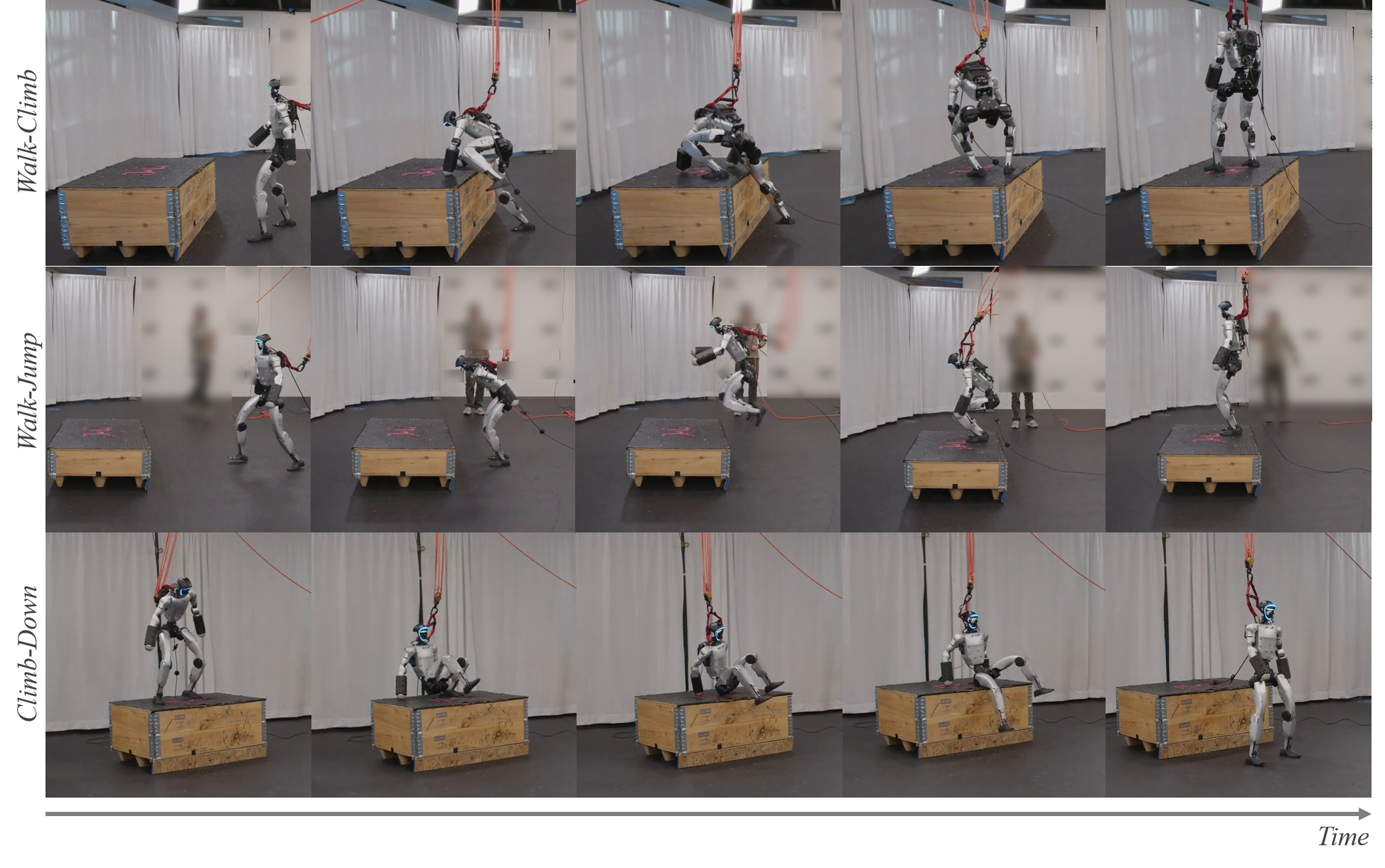}
    \vspace{-0.3in}
    \captionof{figure}{A humanoid robot performs human-like walking, jumping, and climbing behaviors in a box-based environment.
    }
    \vspace{-0.1in}
    \label{fig:teaser}
    \setcounter{figure}{1}
  }
\makeatother

\maketitle

\begin{abstract}
Learning agile humanoid behaviors from human motion offers a powerful route to natural, coordinated control, but existing approaches face a persistent trade-off: reference-tracking policies are often brittle outside the demonstration dataset, while purely task-driven Reinforcement Learning (RL) can achieve adaptability at the cost of motion quality. We introduce a unified multi-task RL framework that bridges this gap by treating reference motion as a prior for behavioral shaping rather than a deployment-time constraint. A single goal-conditioned policy is trained jointly on two tasks that share the same observation and action spaces, but differ in their initialization schemes, command spaces, and reward structures: (i) a reference-guided imitation task in which reference trajectories define dense imitation rewards but are not provided as policy inputs, and (ii) a goal-conditioned generalization task in which goals are sampled independently of any reference and where rewards reflect only task success. By co-optimizing these objectives within a shared formulation, the policy acquires structured, human-like motor skills from dense reference supervision while learning to adapt these skills to novel goals and initial conditions. This is achieved without adversarial objectives, explicit trajectory tracking, phase variables, or reference-dependent inference. We evaluate the method in a challenging box-based parkour playground that demands diverse athletic behaviors (e.g., jumping and climbing), and show that the learned controller transfers beyond the reference distribution while preserving motion naturalness. Finally, we demonstrate long-horizon behavior generation by composing multiple learned skills, illustrating the flexibility of the learned polices in complex scenarios. Results are best visualized through {\url{https://youtu.be/9NamvWhtFPM}.}
\end{abstract}

\IEEEpeerreviewmaketitle

\section{Introduction}

Deep reinforcement learning (DRL) has become a central tool for synthesizing feedback controllers for humanoid robots, enabling policies that can robustly execute task objectives under complex contacts and high-dimensional dynamics. Much of this progress has been driven by goal-oriented formulations, in which policies are trained to directly optimize some task-centric reward. However, achieving complex, contact-rich skills while avoiding unintended emergent behaviors often requires heavy reward shaping and extensive iteration.

In parallel, motion imitation has emerged as a powerful paradigm for learning expressive and coordinated whole-body behaviors by leveraging motion capture data or reconstructed human motion. By providing a strong behavioral prior, imitation-based reinforcement learning can produce agile and human-like skills. However, controllers trained around fixed reference motions can reduce flexibility in steerable or target-driven settings, where the robot must react to new goals, environments, or task conditions, and supplying suitable trajectories across diverse situations becomes impractical.

These two perspectives expose a persistent trade-off: controllers that are tightly guided by reference behaviors can struggle to generalize beyond demonstrated scenarios, while controllers optimized primarily for task objectives may lose motion quality and style. Below, we summarize representative reference-driven RL-based approaches and highlight why closing this gap remains challenging. 

\textbf{Motion tracking.} Early reference-driven approaches rely on motion-tracking objectives that encourage policies to closely follow reference trajectories, enabling stable reproduction of complex whole-body motions in simulation and on real robots~\cite{peng2018deepmimic,zest,he2025asap,chen2025gmt,xie2025kungfubot,weng2025hdmi, xu2025parc}. \emph{DeepMimic}~\cite{peng2018deepmimic} is an early reference-guided RL framework that learns robust physics-based character skills by optimizing an imitation reward over motion clips, producing agile behaviors that can recover from perturbations and can be combined with simple task objectives. More recently, variants of this paradigm have explicitly targeted sim-to-real transfer to deploy such behaviors on physical systems. For instance, \emph{GMT}~\cite{chen2025gmt} scales whole-body motion tracking to large, diverse motion datasets and demonstrates transfer to a physical humanoid (Unitree's G1 robot). \emph{ZEST}~\cite{zest} streamlines sim-to-real motion imitation across heterogeneous reference sources (mocap, monocular video, and animation) using adaptive sampling and an automatic assistive-wrench curriculum, enabling zero-shot deployment of long-horizon, contact-rich skills across multiple robot embodiments. While these methods demonstrate that faithful imitation of human motion data can produce agile, highly coordinated behaviors, the learned controllers remain fundamentally coupled to the demonstrations, limiting their ability to generalize to novel goals, environments, or task conditions.

\textbf{Distillation.} To relax strict tracking and improve adaptability and steerability, several approaches combine motion tracking with distillation~\cite{he2024omnih2o, allshire2025visual,han2025kungfubot2,liao2025beyondmimic,tessler2024maskedmimic}. In these formulations, tracking-based controllers act as teachers, and their behaviors are distilled into policies that no longer require reference motion at execution time, allowing them to be directed by alternative commands or target specifications. Such approaches have been explored in applications including teleoperation~\cite{he2024omnih2o, lu2024pmp}, locomotion~\cite{han2025kungfubot2, ji2024exbody2}, and loco-manipulation~\cite{yin2025visualmimic}. However, because distilled policies inherit their behavior from motion-tracking controllers trained on a fixed set of reference motions, they remain strongly constrained by the reference data used during training, making it difficult to handle out-of-distribution task conditions or environments that differ significantly from the demonstration data. In many such pipelines, the distillation stage primarily provides a strong initialization, but achieving robustness on downstream, task-driven objectives still requires additional RL fine-tuning and task-specific reward design. This results in a multi-stage process---e.g., training a teacher, distilling to a student, and then optionally applying additional RL fine-tuning for a target task---rather than directly optimizing a single policy for the final objective.

\textbf{Adversarial imitation.} Another line of work adopts Adversarial Imitation Learning (AIL) to relax strict trajectory tracking by matching the distributions of generated motion and reference data~\cite{peng2021amp,ren2025humanoid, ma2025styleloco, wang2025more,tang2024humanmimic,pan2025tokenhsi, peng2022ase, wang2024strategy,dou2023c}. These methods typically combine task rewards with imitation rewards learned through discriminators. By operating at the distribution level, AIL approaches allow greater flexibility than step-by-step tracking. However, learning a stable imitation objective becomes particularly challenging in dynamic or contact-rich settings, where small deviations in contact timing and configuration can induce large, discontinuous changes in system behavior. In such regimes, matching motion distributions does not reliably translate to physically valid or robust behavior, and conflicts between task and imitation rewards can undermine both motion quality and task performance.

\textbf{Other formulations.} Beyond these categories, several works explore alternative formulations for incorporating reference motion. SONIC~\cite{luo2025sonic} scales motion tracking to large motion dataset and introduces encoders that map multimodal conditions to kinematic plans. Wen et al.~\cite{wen2025constrained} propose alternating between tracking-based and style-based rewards to trade off task success and motion consistency. Hybrid Imitation Learning (HIL)~\cite{wang2025hil} adopts a similar joint training approach, combining motion tracking with adversarial imitation learning to balance motion fidelity and task performance. However, these adversarial objectives can be difficult to stabilize and tune for hardware. In addition, reference-free tracking is particularly challenging without a proper curriculum.

Overall, most existing reference-driven approaches remain fundamentally constrained by how the reference motion is used: as a trajectory to be followed, a teacher to be distilled, or a distribution to be matched. These formulations make it difficult to decouple motion quality from the specific reference data and to robustly adapt learned behaviors to novel task conditions that fall outside the training distribution.

To address this gap, we propose a unified multi-task learning framework that combines reference imitation with goal-conditioned Reinforcement Learning (RL) to transfer reference-guided behaviors to general task execution. Rather than treating reference tracking as the ultimate objective, we use it as a form of behavioral shaping, providing a strong inductive bias toward natural and coordinated motion. We train a single policy across two tasks that share a common, goal-conditioned observation space. In the reference-guided imitation task, the policy does not receive reference motions as input. Instead, reference motions are used only to define goal conditions and to shape imitation rewards. Consequently, the same policy is trained on a goal-conditioned generalization task in which goals are randomly specified without any reference motion, and rewards are defined purely by task objectives, requiring the policy to adapt its behavior to reach the target. By jointly optimizing both tasks within a shared observation space, the policy acquires motor behaviors that are structured by reference motion yet transferable to novel goals. In particular, this multi-task formulation enables transfer from imitation to general task learning when the downstream objective can be achieved predominantly by reusing---and appropriately sequencing or steering---the motor skills acquired during imitation, rather than requiring fundamentally new behaviors.

We study this question in the context of whole-body mobility over challenging terrain and obstacles. Rough-terrain traversal has become a key testbed for evaluating humanoid motor capabilities, and many recent works have demonstrated impressive adaptability in such complex scenarios~\cite{cui2024adapting, he2025attention,wang2025beamdojo,zhuang2024humanoid,wang2025more,ben2025gallant, gu2024advancing}. However, the resulting behaviors are primarily locomotion-centric, with the core motor behavior remaining walking and adaptation achieved through continuous adjustment of foot placement and body posture. In this work, we focus on extending beyond locomotion-centric terrain traversal toward agile parkour behaviors shaped by human data. While our hardware experiments rely on motion capture (MoCap) for global state (pose) feedback, we believe the same framework can be readily extended to incorporate onboard exteroceptive sensing. To evaluate this capability, we design a box-based parkour playground in which the humanoid must navigate obstacles using a range of athletic skills, including walking, jumping, climbing, and turning. We learn a diverse set of motion skills (as shown in Fig.~\ref{fig:teaser}) and enable their flexible deployment via composition in parkour-style box environments, and we demonstrate long-horizon parkour behaviors by composing skills based on the box layout. This setup highlights the robustness and generalization of the learned policies: the robot can execute extended parkour sequences without careful tuning of initial conditions or task-specific resets, demonstrating reliable deployment across complex, varied environments.

\section{Methodology}

\subsection{Overview}

We study the problem of learning humanoid controllers for challenging dynamic tasks that require both human-like movements and adaptation to changing environments. Specifically, we consider a parkour environment constructed from boxes, where a humanoid must combine skills such as walking, jumping, and climbing to navigate varied obstacle layouts. To address this challenge, we train goal-conditioned policies in a multi-task setup that combines (i) reference-guided imitation and (ii) goal-driven generalization. The resulting policy does not rely on reference motions at inference time; instead, it conditions only on the current state and a goal specification.

In the motion-imitation task, reference motions are used to construct goal conditions and define tracking rewards, providing dense supervision that shapes the policy toward stable and human-like behaviors. In the generalization task, goals are specified independently of reference motion, and rewards are defined purely by task objectives. Jointly optimizing these two tasks allows the policy to acquire natural motor skills through imitation while learning to adapt and extend these skills across diverse conditions.

\subsection{Design Rationale}

The reference-guided imitation task serves purposes beyond enforcing stylistic qualities. In our learning framework, it plays two complementary roles:
\begin{enumerate}
    \item \textbf{Representation learning.} The imitation task encourages the policy to learn a structured mapping between goal conditions and motor behaviors. Because the policy is conditioned only on goal information rather than explicit reference trajectories or phase variables, the resulting behaviors are not tied to individual demonstrations. Instead, the policy learns to represent a family of motor skills, enabling reuse and adaptation.
    \item \textbf{Training stability and efficiency.} The imitation task provides dense reward signals that directly supervise how the humanoid should move, encoding rich information about pose coordination and timing derived from human motion data. These dense, human demonstration-based rewards provide stronger, more informative learning signals than task-level objectives alone, enabling the policy to efficiently acquire high-quality motor skills.
\end{enumerate}

Overall, our formulation yields a single goal-conditioned policy that combines natural motion with goal-driven adaptability, without relying on adversarial discriminators, explicit trajectory tracking, or reference motions at inference time.

\subsection{Training Setup and MDP Formulation}

In our goal-conditioned RL framework, both the imitation and the generalization tasks are formulated in a unified setting that shares the same policy parameters, observation, and action spaces. The two tasks differ only in how goals, rewards, and value estimation are defined. The following presents the Markov Decision Process (MDP) formulation and the training setup of the proposed method. 

\textbf{Observation and goal representation.}
At each timestep, the policy observes the robot state $s_t$, which includes the humanoid's joint configuration and velocity, projected gravity, and torso angular velocities in a character-centric coordinate frame. The policy is also conditioned on a goal variable $g_t$ that specifies the motion's target. Importantly, the policy does not observe reference motions, future trajectories, or explicit phase variables. All task-specific information is conveyed through the goal condition, ensuring a consistent observation space across training tasks. The goal variable $g_t$ is a target root (torso) location in the horizontal plane, represented as a 2D position $(x, y)$ relative to the character. In the imitation task, the goal, $g_t$, is derived from the references. In the generalization task, the goal is randomly sampled, independently of any reference motion.

\textbf{Action space.}
The humanoid is actuated using joint-level Proportional-Derivative (PD) controllers. The policy outputs residual actions $\bm{a}_t$, which are mapped to commanded joint position targets and converted to joint torques:
\[
\bm{q}^{\mathrm{cmd}}_j = \bar{\bm{q}} + \bm{\Sigma}\,\bm{a}_t,
\]
where $\bar{\bm{q}}$ denotes a set of default joint positions and $\bm{\Sigma}$ is a positive-definite diagonal matrix of per-DoF action scales. The PD gains are specified according to the procedure in~\cite{zest}, modeling each joint as an independent second-order system. Unlike standard motion-tracking formulations that use reference joint positions as a feedforward term in the command to stabilize and accelerate learning, we omit this term to preserve the ability to deviate from the reference when transitioning to task-driven objectives.

\textbf{Reward design.}
The imitation task and the generalization task differ in their reward definitions. In the imitation task, the reward encourages the character to match the reference motion, providing dense supervision on how natural human motion should be executed. The total imitation reward consists of three components: a tracking reward, a regularization reward, and a survival reward:
\begin{equation}
    r^{\text{imi}}_t \;=\; r_{\text{track},t} \;+\; r_{\text{reg},t} \;+\; r_{\text{surv},t}.
\end{equation}

The tracking reward encourages the policy to follow the reference motion and provides dense behavioral shaping. It measures the discrepancy between the current state and the reference in terms of base pose, base velocity, and joint configurations:
\begin{equation}
    r_{\text{track}} 
    \;=\; 
    \sum_{i} c_{t_i} \exp\!\Bigl(-\,\kappa\frac{\|\mathbf{e}_i\|^2}{\sigma_i^2}\Bigr),
\end{equation}
where $\mathbf{e}_i$ denotes the tracking error for term $i$, $\sigma_i$ is a normalization scale, $c_{t_i}$ is a weighting coefficient, and $\kappa$ is a temperature parameter. The regularization reward aggregates penalties that promote physically plausible and smooth behavior, including action smoothness, joint torque usage, and violations of joint limits. The survival reward is a constant positive term provided at each timestep to discourage premature termination and encourage longer, stable rollouts.

For the generalization task, the reward no longer depends on reference motion and instead combines a sparse goal-driven task reward with the same regularization and survival terms as in the imitation task:
\begin{equation}
    r^{\text{gen}}_t \;=\; r_{\text{goal},t} \;+\; r_{\text{reg},t} \;+\; r_{\text{surv},t}.
\end{equation}

The goal-driven reward encourages progress toward and achievement of task-specific objectives, such as reaching a target base position and orientation, and is intentionally sparse compared to the imitation tracking reward. In general, we instantiate it as a weighted combination of progress and completion terms:

\begin{equation}
r_{\text{goal},t}
=
- c_p \left\| \mathbf{e}^{xy}_t \right\|_2^2
- c_o \left\| \boldsymbol{\theta}_t \right\|_2^2
+ r_{\text{reach},t} ,
\end{equation}
where \( \mathbf{e}^{xy}_t = \mathbf{p}^{xy}_t - \mathbf{p}^{xy}_{goal} \) denotes the base position distance in the horizontal plane, and
\( \boldsymbol{\theta}_t = \mathrm{AxisAngle}\!\left(\boldsymbol{\Phi}_t \otimes (\boldsymbol{\Phi}_{goal})^{-1}\right) \) represents the base orientation error computed from the axis--angle magnitude of the quaternion difference between the current and target base orientations. The weights \( c_p \) and \( c_o \) control the relative weights. The constant reward term $r_{\text{reach},t}$ is activated only when the robot reaches the goal. 

As a result, the policy relies on behaviors shaped during imitation and adapts them to satisfy the task objectives under varying conditions. The regularization and survival rewards serve the same role as in the imitation task. Further details regarding the rewards are provided in the appendix.

\textbf{Curriculum}
Building on prior work on assistive-wrench curricula for RL-based motion tracking~\cite{zest}, we employ an automatic difficulty schedule that stabilizes early-stage learning and enables reliable acquisition of highly dynamic behaviors. Concretely, we maintain a global scalar difficulty variable $\lambda \in [0,1]$ that is adapted online based on training performance: $\lambda$ increases when the agent (i) avoids early termination and (ii) remains \emph{roughly} consistent with the reference under a relaxed tracking criterion, and decreases otherwise. This single scalar controls both (a) the magnitude of a virtual spatial assistive wrench applied at the base and (b) the probability of sampling the reference-guided imitation task versus the goal-conditioned generalization task. In our experiments, this coupled curriculum was crucial for speeding up convergence—especially for box-climbing behaviors—by reducing catastrophic failures early on via transient physical assistance (which is particularly beneficial when a reference feedforward command term such as $\bm{q}^{\mathrm{ref}}$ is absent from the action parametrization), while simultaneously orchestrating a smooth shift from dense, reference-based shaping to sparser, goal-driven supervision by progressively reweighting the sampling of imitation and generalization tasks as training stabilizes.

\smallskip
\noindent\emph{Virtual wrench computation.}
Let $(\mathbf{p},\mathbf{v},\boldsymbol{\Phi},\boldsymbol{\omega})$ denote the base position, linear velocity, orientation, and angular velocity, and let $(\hat{\mathbf{p}},\hat{\mathbf{v}},\hat{\boldsymbol{\Phi}},\hat{\boldsymbol{\omega}})$ denote their reference counterparts. We compute a nominal spatial wrench at the base using a PD term on the base pose tracking error, together with a feedforward component that compensates nominal torso dynamics:
\begin{subequations}\label{eq:assist}
\begin{align}
\mathbf{F}_{b} \;&=\; M\!\left(\hat{\dot{\mathbf{v}}}
\;+\; k_p^v\,(\hat{\mathbf{p}}-\mathbf{p})
\;+\; k_d^v\,(\hat{\mathbf{v}}-\mathbf{v})
\;-\; \mathbf{g}\right), \label{eq:assist-F}\\[2pt]
\mathbf{M}_{b} \;&=\;
\begin{aligned}[t]
\mathbf{I}\,\hat{\dot{\boldsymbol{\omega}}}
&\;+\; k_p^\omega\,\mathbf{I}\,\bigl(\hat{\boldsymbol{\Phi}}\boxminus \boldsymbol{\Phi}\bigr)
\;+\; k_d^\omega\,\mathbf{I}\,(\hat{\boldsymbol{\omega}}-\boldsymbol{\omega}) \\
&\;+\; \boldsymbol{\omega}\times(\mathbf{I}\boldsymbol{\omega})
\;-\; \mathbf{r}_{\mathrm{b,com}}\times M\mathbf{g},
\end{aligned}
\label{eq:assist-M}
\end{align}
\end{subequations}
where $M$ and $\mathbf{I}$ are the whole-body mass and nominal base inertia at a default configuration, $\mathbf{g}$ is gravity, $\mathbf{r}_{\mathrm{b,com}}$ is the position of the whole-body CoM with respect to the base, and $\boxminus$ denotes the Lie-group difference on $SO(3)$. The applied assistive wrench is
\[
\mathbf{w}_e \;=\; \beta(\lambda)\begin{bmatrix}\mathbf{F}_b \\ \mathbf{M}_b\end{bmatrix},
\qquad
\beta(\lambda) \;=\; (1-\lambda)\,\beta_{\max}, \ \ \beta_{\max}<1,
\]
so assistance is strong at low difficulty ($\lambda \!\approx\! 0$) and vanishes as training progresses ($\lambda \!\to\! 1$), while remaining partial.

\smallskip
\noindent\emph{Task mixing.}
The same scalar $\lambda$ governs the transition from pure imitation to mixed training via a linear interpolation of the imitation sampling probability:
\[
p_{\text{imi}}(\lambda) \;=\; (1-\lambda)\,p_0 \;+\; \lambda\,p_{\text{target}}, \ \ p_0 > p_{\text{target}},
\]
with $p_0$ and $p_{\text{target}}$ denoting the imitation-task sampling probabilities at the lowest and highest difficulty, respectively. In parallel, we expand the range of training configurations (initial states and task goals) as $\lambda$ increases. Overall, this curriculum allows the policy to first acquire structured, coordinated motor skills under strong stylistic regularization and partial physical assistance, and then progressively adapt these skills to broader task conditions as external support is removed.

\textbf{State initialization and randomization.}
State initialization plays a critical role in our framework, as it guides how skills shaped in the motion-imitation task are transferred and generalized in the generalization task. We adopt different state and goal initialization strategies for the motion-tracking and generalization tasks. In the motion-tracking task, the humanoid's state and goal are initialized by sampling from reference motions with only small perturbations. This initialization strategy anchors the policy in reference-like configurations, ensuring that learning focuses on acquiring high-quality motor skills rather than executing in novel situations. In contrast, the generalization task adopts a much broader initialization strategy. Both the humanoid state and the goal are sampled from wide distributions, similar to standard RL settings. States are drawn from a more diverse set of configurations, and goals are specified randomly across different locations. This initialization strategy exposes the policy to conditions that deviate from the reference data, requiring it to adapt to novel goals and environments.

To further improve robustness on hardware, we apply domain randomization to some simulation parameters during training, including contact friction coefficients and link masses. Additionally, occasional external pushes are introduced to the torso, and noise is added to observations. These perturbations expose the policy to different dynamics and sensing conditions, helping it learn behaviors that remain stable when transferred to the hardware.

\textbf{Training setup.}
Training is performed in Isaac Lab \cite{mittal2025isaac} utilizing Proximal Policy Optimization (PPO) \cite{ppo} as the RL algorithm. We adopt an asymmetric actor-critic architecture \cite{AsymmetricActorCritic}. The policy (actor) is parameterized as a three-layer MLP that maps the state $(s_t, g_t)$ to a Gaussian action distribution over PD target joint positions. The value function (critic) is parameterized as a three-layer MLP, but is also provided with an additional task indicator variable $k_t$ that specifies whether the current sample comes from the motion imitation task or the generalization task. In addition, the critic receives privileged information available in simulation but not in the real world, such as the full root state, contact forces, and the assistive wrench signal. This privileged input is used solely for value estimation to improve training stability and sample efficiency.

\begin{figure}[t]
    \centering
    \begin{subfigure}{0.24\textwidth}
        \centering
        \includegraphics[width=1.0\columnwidth]{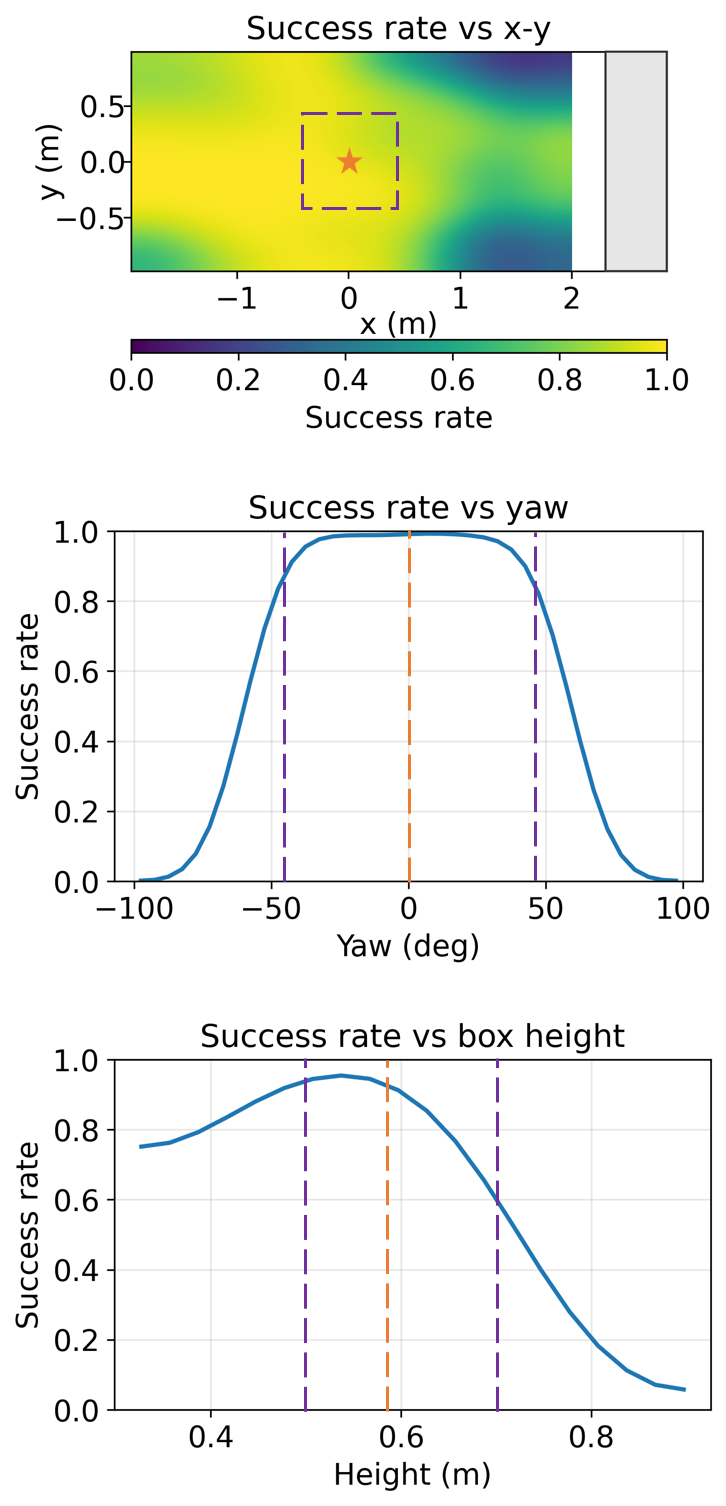}
        \caption{Walk-Climb}
    \end{subfigure}
    \begin{subfigure}{0.24\textwidth}
        \centering
        \includegraphics[width=1.0\columnwidth]{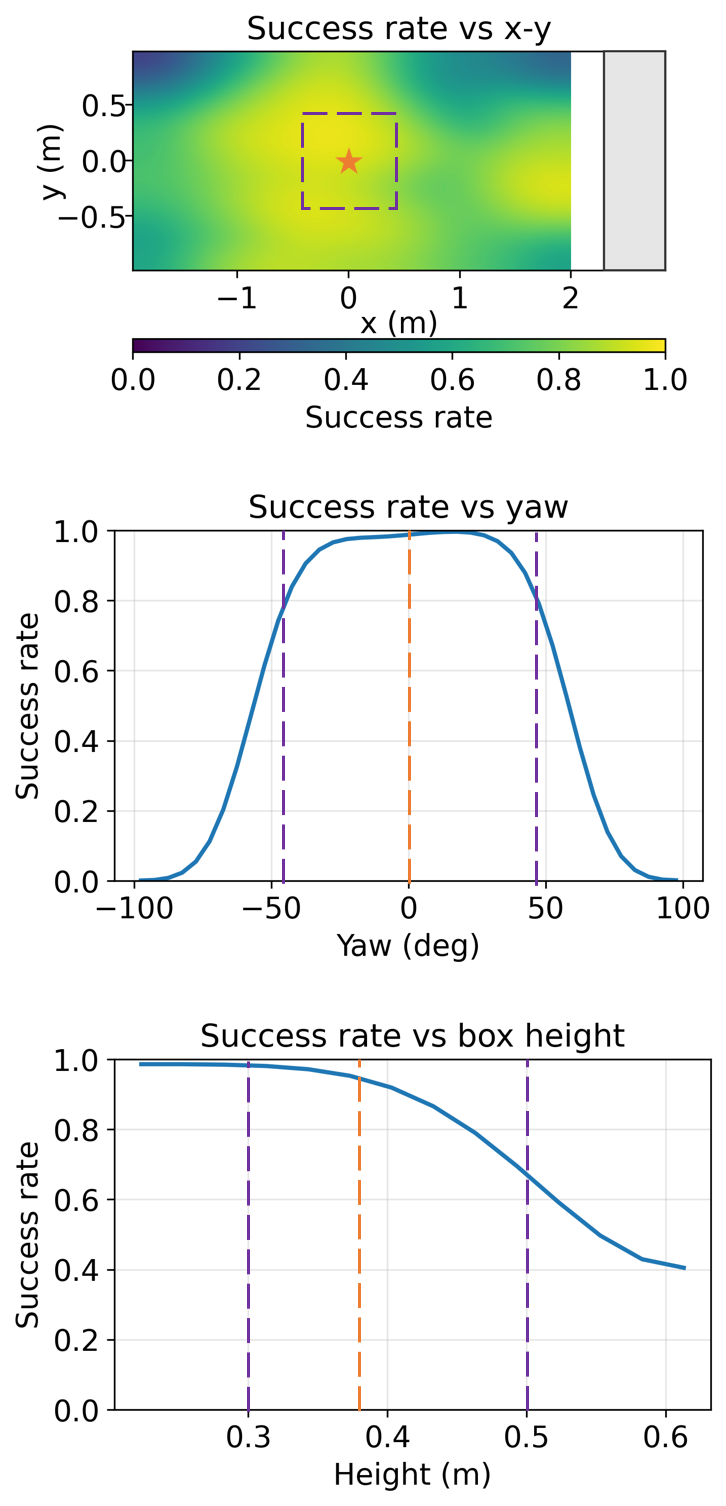}
        \caption{Walk-Jump}
    \end{subfigure}
    \caption{Success rate of our method under different initial conditions for walk-climb and walk-jump skills. When varying one initial condition, all other conditions are held at their nominal values. Orange markers show the nominal configuration of the initial state, while the purple markers show the randomness level of the initialization during training. The gray rectangle represents the box with its edge positioned at 2.3m.}
    \vspace{-0.in}\label{fig:init_condition}
\end{figure}

\begin{figure}[t]
    \centering
\includegraphics[width=1.0\columnwidth]{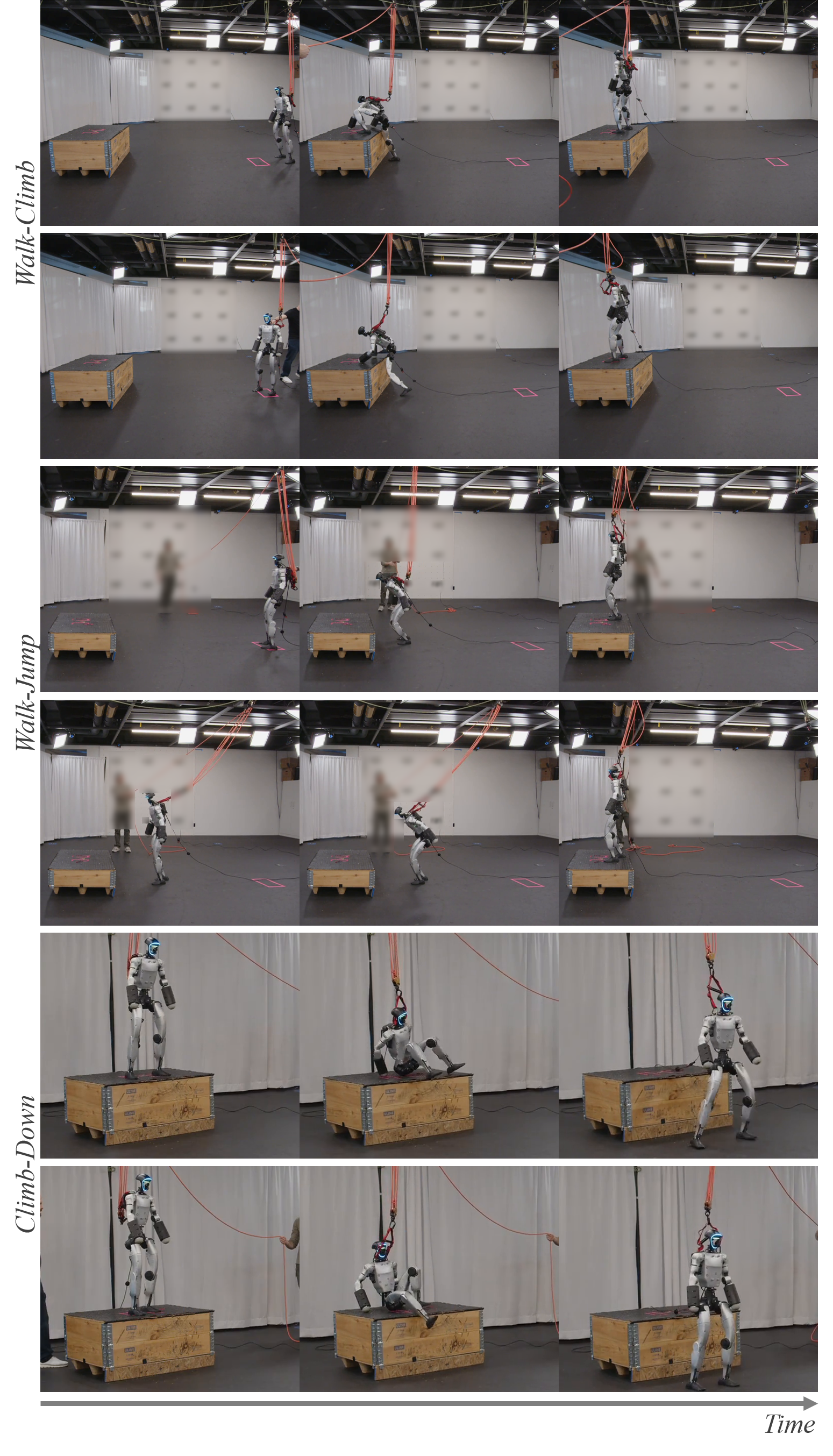}
    \vspace{-0.25in}
    \caption{Hardware experiments with varied initial conditions for the walk–climb, walk–jump, and climb–down skills. Each behavior is depicted from two different initial conditions. Despite changes in initial conditions, the robot adapts its strategy and successfully executes the skills.}
    \vspace{-0.2in}
    \label{fig:init_condition_hardware}
\end{figure}



\section{Evaluation}
In this section, we conduct experiments both in simulation and on hardware to answer the following questions:

\begin{itemize}
    \item \emph{\textbf{Robustness and Generalization in Simulation and on Hardware:} How robust are the learned policies under nominal and beyond-nominal conditions, and how well do they generalize to variations in initial states and task configurations in both simulation and real-world experiments?}
    \item \emph{\textbf{Comparison to Alternative Training Paradigms:} How does our method perform overall compared to tabula rasa (pure) RL, tracking-based RL in terms of task success, robustness, and motion quality?}
    \item \emph{\textbf{Long-Horizon Skill Composition:} Can the learned policies be reused as modular skills and composed into sequences to solve long-horizon parkour scenarios?}
    \item \emph{\textbf{Key Components and Ablations:} Which components are essential for the pipeline to work effectively, as determined through simulation-based ablation studies?}
\end{itemize}

All experiments are performed using the Unitree G1 humanoid ($29$ DoF, $1.2\,\mathrm{m}$ tall, $35\,\mathrm{kg}$), with simulation evaluations conducted either in MuJoCo \cite{todorov2012mujoco} or Isaac Lab \cite{mittal2025isaac}. We evaluate three representative behaviors: walk-jump, walk-climb, and climb-down. We train one policy per behavior using a single reference motion, with the walk–climb and walk–jump behaviors augmented by their mirrored motions.


\subsection{How Robust and Generalizable Are the Learned Policies in Simulation and on Hardware?}
\label{subsec: evaluation_a}
We assess robustness and generalization by introducing controlled variability in initial conditions, and measure how reliably the learned behaviors extend beyond the nominal training configuration while preserving motion quality. Specifically, we report task success rates under systematic variations of the initial root position, heading orientation, and box height, using the walk-climb and walk-jump tasks as representative examples. We evaluate a thousand trials for each task and a trial is considered successful if the robot ends in a stable standing configuration on top of the box, with the base height within $10,\mathrm{cm}$ of the target height ($0.8,\mathrm{m}$), and the base position within $20,\mathrm{cm}$ of the box center in the horizontal plane (the box half-width from edge to center is $40,\mathrm{cm}$). We focus on walk–climb and walk–jump because these tasks naturally involve a walking phase, which allows the robot to approach the box from a wide range of initial positions and thus provides a meaningful test of generalization under beyond-nominal conditions. As shown in Fig.~\ref{fig:init_condition}, our method maintains high success rates across a wide range of perturbations to initial conditions, demonstrating that the learned behaviors are not narrowly tied to specific reference executions but instead generalize well beyond the nominal starting state.


To provide a qualitative illustration of the policy's generalization capability, we conduct a sim-to-real evaluation on hardware, as shown in Fig.~\ref{fig:init_condition_hardware}. The figure includes all three representative behaviors: walk-climb, walk-jump, and climb-down. 
For the walk-climb skill, the robot is initialized at different distances from the box, ranging from far to near. The robot is able to walk toward the box and successfully climb onto it. Notably, the policy naturally adapts its strategy by leading with either the left or the right leg, depending on the initial configuration and dynamics. For the walk-jump skill, we observe similar adaptive behavior. When the robot starts farther from the box, it first walks forward, then jumps. In contrast, when the initial distance is sufficiently small, the robot directly initiates a jump without any walking motion. For the climb-down skill, the robot also exhibits diverse strategies under different initial conditions. In particular, we observe cases in which the robot repeatedly uses one foot (e.g., the left foot) to push against the box, gradually shifting its center of mass forward before stepping down. These results demonstrate that the learned policies do not simply replay a single reference trajectory, but instead generalize effectively to different task configurations.

\begin{table*}[t]
    \centering
    \caption{Quantitative comparison with alternative training paradigms. Task success rate reflects robustness, while base orientation (root $R$) and joint position (joint pos $q$) errors measure similarity to the reference motion. The latter metrics are omitted for tabula rasa RL, which does not use reference information. Arrows in the table indicate the desired direction (increase or decrease) for each metric.}
    \label{tab:baselines}
    \setlength{\tabcolsep}{4pt}
    \renewcommand{\arraystretch}{0.95}
    \begin{tabular}{lccccccccc}
        \toprule
        \textbf{Methods} 
        & \multicolumn{3}{c}{\textbf{Walk-Jump}} 
        & \multicolumn{3}{c}{\textbf{Walk-Climb}} 
        & \multicolumn{3}{c}{\textbf{Climb-Down}} \\
        \cmidrule(lr){2-4}
        \cmidrule(lr){5-7}
        \cmidrule(lr){8-10}
        & success rate $\uparrow$ & root $R$ $\downarrow$ & joint pos $q$ $\downarrow$
         & success rate $\uparrow$ & root $R$ $\downarrow$ & joint pos $q$ $\downarrow$
         & success rate $\uparrow$ & root $R$ $\downarrow$ & joint pos $q$ $\downarrow$ \\
        \midrule

        \multicolumn{10}{l}{\emph{Nominal
Initializations}} \\
        \midrule
        ZEST mocap & 0.98 & 0.45 & 1.60 & 0.92 & 0.83 & 1.60 & 0.90 & 1.02 & 1.75  \\
        Tabula rasa RL & 1.00 & - & - & 1.00 & - & - & 0.91 & - & - \\
        Ours & 1.00 & 0.32  & 2.26  & 1.00  & 0.31 & 1.87  & 1.00 & 0.73 & 3.43    \\
        \midrule
        \multicolumn{10}{l}{\emph{Beyond Nominal
Initializations}} \\
\midrule
        ZEST mocap & 0.17 & 0.65 & 2.15 & 0.57 & 1.44 & 2.87 & 0.90 & 1.04 & 1.81 \\
        Tabula rasa RL & 0.54 & - & - & 0.53 & - & -  & 0.91 & - & - \\
        Ours & 0.62 & 0.99 & 3.34  & 0.76  & 1.41 & 4.41 & 0.98 & 0.74 & 3.44    \\
      
        \bottomrule
    \end{tabular}
\end{table*}

\begin{figure*}[t]
    \centering
    \includegraphics[width=\textwidth]{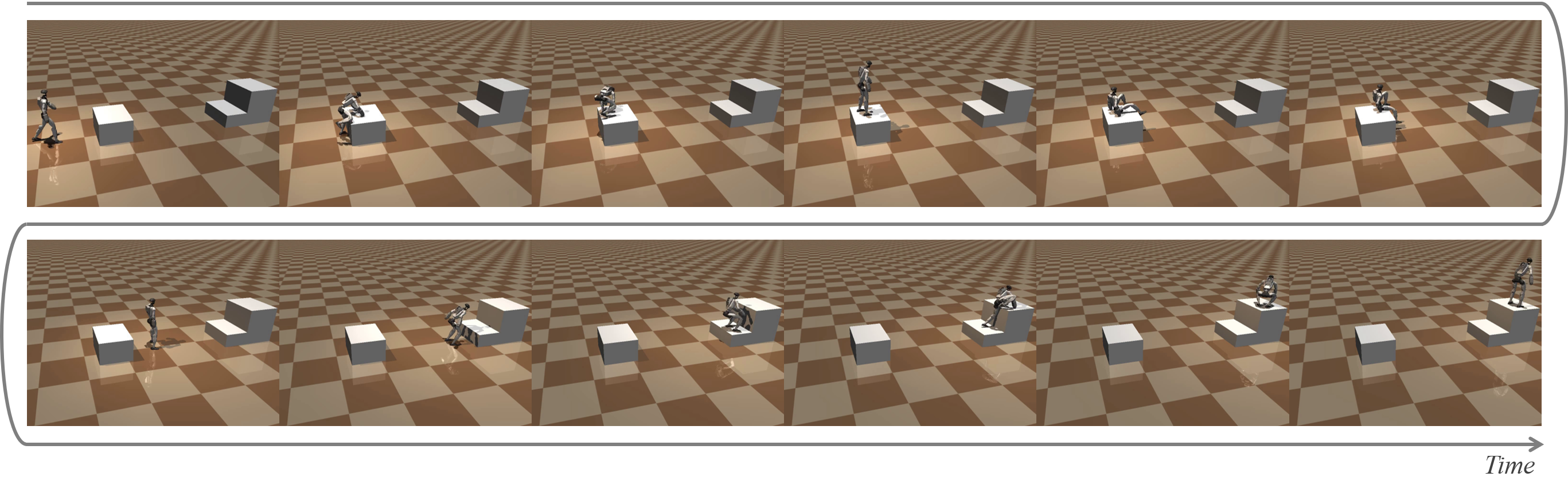}
    \vspace{-0.3in}
    \caption{Multi-skill composition in a sim-to-sim evaluation in MuJoCo. Learned policies are composed to execute walk-climb, walk-jump, and climb-down behaviors over long horizons.}
    \vspace{-0.1in}
    \label{fig:compose}
\end{figure*}



\begin{table*}[t]
    \centering
    \caption{Quantitative comparison with ablative baselines. Task success rate reflects robustness, while base orientation (root $R$) and joint position (joint pos $q$) errors measure similarity to the reference motion. Arrows in the table indicate the desired direction (increase or decrease) for each metric.}
    \label{tab:ablation}
    \setlength{\tabcolsep}{4pt}
    \renewcommand{\arraystretch}{0.95}
    \begin{tabular}{lccccccccc}
        \toprule
        \textbf{Methods} 
        & \multicolumn{3}{c}{\textbf{Walk-Jump}} 
        & \multicolumn{3}{c}{\textbf{Walk-Climb}} 
        & \multicolumn{3}{c}{\textbf{Climb-Down}} \\
        \cmidrule(lr){2-4}
        \cmidrule(lr){5-7}
        \cmidrule(lr){8-10}
        & success rate $\uparrow$ & root $R$ $\downarrow$ & joint pos $q$ $\downarrow$
         & success rate $\uparrow$ & root $R$ $\downarrow$ & joint pos $q$ $\downarrow$
         & success rate $\uparrow$ & root $R$ $\downarrow$ & joint pos $q$ $\downarrow$ \\
        \midrule

        \multicolumn{10}{l}{\emph{Nominal
Initializations}} \\
        \midrule
        Ours w/o task curriculum & 0.00 & 0.38 & 2.71 & 0.00 & 0.51 & 2.70 & 0.96 & 0.58 & 2.83 \\
       
        Ours w/o imitation & 0.00 & 4.52  & 11.3  & 0.00 & 2.88 & 8.00  & 0.98 & 1.88 & 16.12 \\

         Ours w/o generalization & 0.99 & 0.29 & 2.25   & 0.97 & 0.31 & 1.99  & 0.91 & 0.56 & 2.56 \\
         
        Ours & 1.00 & 0.32  & 2.26  & 1.00  & 0.31 & 1.87  & 1.00 & 0.73 & 3.43 \\
        \midrule
        \multicolumn{10}{l}{\emph{Beyond Nominal
Initializations}} \\
\midrule
  Ours w/o task curriculum & 0.00 & 1.01 & 3.34  & 0.00 &1.20  &  4.04 & 0.69 & 1.01 & 3.81  \\
        
        Ours w/o imitation & 0.00 & 4.75  & 12.97 & 0.00 & 3.57  & 10.09  & 0.97 & 3.75 & 16.30 \\

        Ours w/o generalization & 0.27 & 2.10 & 4.95  & 0.53 & 1.78  &  5.06  & 0.70 & 0.79 & 3.30 \\

        Ours & 0.62 & 0.99 & 3.34  & 0.76  & 1.41 & 4.41 & 0.98 & 0.74 & 3.44 \\
      
        \bottomrule
    \end{tabular}
    \vspace{-0.1in}
\end{table*}

\subsection{How Does Our Method Compare to Tabula Rasa RL and Pure Motion Imitation?}
We compare our method against two representative baselines: a tracking-based RL approach (\emph{MoCap-based ZEST}~\cite{zest}) and \emph{tabula rasa RL}. We augment ZEST with motion capture data to get the relative position and orientation between the robot and the box, and directly track reference motions during execution. In contrast, while tabula rasa RL also takes motion capture data as input, it is trained purely with carefully handcrafted task rewards, without any reference motion or imitation signals.

We evaluate these methods using three metrics: \emph{task success rate} (as defined in Section \ref{subsec: evaluation_a}), \emph{root orientation error}, and \emph{joint position error}. The root orientation error measures the deviation of the projected gravity direction and captures overall balance and stability, while the joint position error reflects motion naturalness. We consider both nominal and beyond-nominal initializations, where beyond-nominal conditions introduce randomized offsets to test robustness. For walk–jump and walk–climb, the initial conditions are perturbed by up to $\pm 2,\mathrm{m}$ forward, $\pm 1,\mathrm{m}$ laterally, and $\pm 45^\circ$ in yaw. For climb-down, where the robot starts on top of the box, smaller perturbations are applied ($-0.3,\mathrm{m}$ to $+0.1,\mathrm{m}$ forward, $\pm 0.2,\mathrm{m}$ laterally, and $\pm 30^\circ$ in yaw).

Quantitative results are summarized in Table~\ref{tab:baselines}.
Across all tasks and conditions, our method achieves the highest success rates under both nominal and out-of-distribution (beyond-nominal) initializations, while maintaining low root orientation error and reasonable joint position error. This indicates that the learned behaviors are both stable and natural, and that they generalize effectively to initial conditions that differ significantly from those seen during training.

ZEST mocap achieves low joint position error under nominal conditions, as expected, since the reference joint positions are explicitly provided as inputs. However, its success rate drops substantially under beyond-nominal initializations. We find that this failure mode is often caused by a strong bias toward tracking the reference motion. For example, when the robot starts closer to the box or ahead of the reference timing, the policy attempts to jump backward abruptly to re-align with the reference trajectory, leading to motion instability. As a result, while joint-level tracking remains accurate when execution succeeds, overall robustness is limited.

The tabula rasa RL policy, on the other hand, exhibits less structured and less natural behavior. Although it can succeed under nominal conditions, its motions are noticeably uncoordinated and aggressive. For example, in the climb-down task, the policy directly jumps off a box of approximately $50\,\mathrm{cm}$ height instead of executing a controlled descent. In the walk-jump and walk-climb tasks, the policy also tends to move aggressively toward the goal, often completing the entire task in roughly $2\,\mathrm{s}$, whereas the corresponding reference motions usually take $10\,\mathrm{s}$. These behaviors reflect the aggressive, greedy nature of the tabula rasa RL policy, which primarily exploits large task rewards without developing robust, coordinated strategies or structured motion. As a result, when evaluated under beyond-nominal initializations, the failure rate of tabula rasa RL increases substantially.

\subsection{Can the Learned Skills Be Composed to Solve Long-Horizon Parkour Scenarios?}
We examine whether the learned policies can be reused as modular skills and composed into longer sequences to solve long-horizon box-parkour scenarios. To this end, we compose skills using a rule-based state machine that issues task-level goals based on the box layout. Fig.~\ref{fig:compose} shows representative long-horizon parkour executions composed from the learned skills, evaluated in MuJoCo, demonstrating sim-to-sim transfer to a different physics engine. In this example, the robot combines walking with climbing up and down, jumping, followed by climbing up. Importantly, these behaviors remain reliable across different initial configurations and target goal specifications, without careful tuning or task-specific resets. Together, these results demonstrate that the learned skills generalize robustly and can be flexibly reused in long-horizon parkour scenarios.

\subsection{Which Components Are Necessary for the Pipeline to Work?}
We perform simulation-based ablation studies to isolate the contributions of key design choices in our pipeline and identify which components are essential for achieving robust, generalizable, and natural behaviors. In particular, we study the roles of the task-mixing curriculum, the reference-guided imitation task, and the goal-conditioned generalization task by removing each component in turn. Quantitative results are summarized in Table~\ref{tab:ablation}.

When the task mixing curriculum is removed, the policy is trained by jointly optimizing the imitation and generalization tasks from the beginning (``Ours w/o task curriculum''). The resulting policy fails completely on walk-jump and walk-climb skills. While the policy is able to learn basic locomotion behaviors such as walking, it consistently fails to acquire more dynamic, contact-rich skills such as jumping and climbing. This result indicates that directly optimizing imitation and generalization objectives is challenging for complex behaviors. Instead, using task curriculum with more imitation for behavior shaping, followed by more generalization task, provides a more effective learning process.

Removing the imitation task (``Ours w/o imitation task'') significantly degrades both task success and motion quality. Because the goal-conditioned reward is simple and intentionally sparse, depending on the 2D distance to the goal and a final reach reward, the policy lacks sufficient guidance to discover coordinated, contact-rich whole-body behaviors. As a result, the learned behavior becomes highly greedy: the robot tends to rush directly toward the goal but fails to exploit structured interactions such as stepping, jumping, and climbing to get onto the box. This variant consistently fails on walk-jump and walk-climb tasks and exhibits substantially worse motion quality, as reflected by large root orientation and joint position errors. For the climb-down task, the policy often resorts to simply jumping straight down. These results demonstrate that sparse goal rewards alone are insufficient for learning agile, coordinated motor skills.

Removing the generalization task (``Ours w/o generalization task'') preserves high success rates and stylistic motion quality under nominal initializations. However, performance degrades sharply under beyond-nominal conditions, with significant drops in success rates for walk-jump and walk-climb. This highlights that imitation alone is insufficient to support robust adaptation to novel task conditions.



\section{Conclusion}
We presented a unified multi-task RL framework for acquiring agile, natural, and deployable humanoid motor skills. The key idea is to treat reference motion as a prior for \emph{behavioral shaping} rather than a deployment-time constraint: a single goal-conditioned policy is trained jointly on a reference-guided imitation task that provides dense supervision and a reference-free generalization task that optimizes task success under diverse initial conditions and commands. Because reference trajectories are never used as policy inputs, the resulting controller executes purely from state and goal, enabling generalization beyond the demonstration dataset. We evaluated the approach via challenging box-based parkour tasks that require a range of athletic behaviors, including jumping and climbing, and showed that multi-task training yields robust transfer while preserving motion naturalness. Finally, we demonstrated long-horizon skill composition using a simple state-machine composer that produces task-level goals to sequence and deploy the learned skills to accomplish parkour objectives.

This work opens several directions for future research. Promising extensions include integrating perception and scene understanding into the high-level composer, enriching the goal interface beyond root targets and discrete behavior labels, and scaling the skill library to more complex contact-rich behaviors. Another key next step is to replace the hand-designed state machine with a learned high-level policy that selects and parametrizes goals for long-horizon decision making. Ultimately, we believe this framework provides a practical pathway from reference-guided skill acquisition to flexible, goal-driven humanoid autonomy in unstructured environments.
\label{sec:conclusion}

\section*{Acknowledgments}
We thank Zach Nobles, Francesco Iacobelli, Scott Biddlestone, Jonathan Foster, Ashley Dodge, and Sylvain Bertrand for their invaluable assistance with the hardware experiments and their support in developing the software infrastructure required for the tests.

\bibliographystyle{plainnat}
\bibliography{references}

\clearpage
\appendix
This appendix provides detailed implementation descriptions of the proposed framework.
We describe the observation design, goal representation, reward formulation, curriculum schedules, network architecture, RL training, state initialization, and domain randomization used in all experiments, with an emphasis on faithful reproducibility.

In addition, we provide a supplementary video that presents extensive qualitative results.
The video includes hardware experiments demonstrating successful execution and generalization of learned behaviors, as well as side-by-side comparisons with baseline methods in simulation under varied initial conditions and environment configurations.

\section{Observations and Goal Representation}
\label{sec:sup_obs_goal}

We adopt an asymmetric actor--critic formulation.
The policy receives only deployment-available proprioceptive observations and a low-dimensional goal command, while the critic is additionally provided with privileged information available in simulation to improve value estimation.
Zero-mean Gaussian noise is injected into selected policy observations to improve robustness.
Unless otherwise noted, we do not apply per-term scaling or clipping. The observation is summarized in Table~\ref{tab:obs_terms}

\begin{table*}[t]
\centering
\small
\caption{Observation terms summary. Noise is zero-mean Gaussian and additive. Privileged observations are used by the critic only.}
\label{tab:obs_terms}
\begin{tabular}{l l l}
\toprule
\textbf{Term Name} & \textbf{Definition} & \textbf{Noise} \\
\midrule
\multicolumn{3}{c}{\textbf{Policy Observations}} \\
\midrule
Torso angular velocity
& $^{T}\omega_{IT}$ (IMU on torso)
& $\mathcal{N}(0,\;0.10^2)$ \\

Projected gravity
& $^{T}g_{I}$ (gravity direction expressed in torso/IMU frame)
& $\mathcal{N}(0,\;0.015^2)$ \\

Joint positions (relative)
& $q_{\text{rel}}$
& $\mathcal{N}(0,\;0.005^2)$ \\

Joint velocities (relative)
& $\dot{q}_{\text{rel}}$
& $\mathcal{N}(0,\;0.25^2)$ \\

Previous action
& $a_{t-1}$
& -- \\

Target command (relative pose)
& $(\Delta x_{\text{goal}},\,\Delta y_{\text{goal}},\,q_{\text{goal}})$
& $\mathcal{N}(0,\;0.015^2)$ \\

\midrule
\multicolumn{3}{c}{\textbf{Privileged Observations (critic only)}} \\
\midrule
Projected gravity (base frame)
& $^{B}g_{I}$
& -- \\

Base linear velocity (base frame)
& $^{B}v_{IB}$
& -- \\

Base angular velocity (base frame)
& $^{B}\omega_{IB}$
& -- \\

Base height
& $^{I}r^{z}_{IB}$
& -- \\

End-effector incoming wrenches
& $\{\,^{B}w_{\text{ee}}\,\}$ (contact wrenches on selected bodies)
& -- \\

End-effector positions w.r.t.\ base
& $\{\,^{B}r_{B\text{ee}}\,\}$
& -- \\

End-effector linear velocities (base frame)
& $\{\,^{B}v_{I\text{ee}}\,\}$
& -- \\

Assistive wrench (fictitious force)
& $f_{\text{assist}}$
& -- \\

Assistive wrench (fictitious torque)
& $\tau_{\text{assist}}$
& -- \\

Wrench scale
& $\beta$
& -- \\

Similarity metric
& $\hat{s}$ (reference-based similarity / tracking score)
& -- \\

Task binary indicator
& $k_t \in \{0,1\}$ (imitation vs.\ generalization)
& -- \\

Reference look-ahead joint delta
& $q^{\ast}_{t+1}-q_t$ (relative reference, look-ahead $=1$)
& -- \\

\bottomrule
\end{tabular}
\end{table*}

\paragraph{Goal representation.}
At each timestep \(t\), the policy receives the proprioceptive state \(s_t\) and a goal variable \(g_t\).
In all box-based tasks, we instantiate the goal as a target \emph{root pose} expressed in the character-centric frame:
\begin{equation}
g_t \;=\; \bigl(\Delta x_{\text{goal}},\, \Delta y_{\text{goal}},\, q_{\text{goal}}\bigr),
\end{equation}
where \((\Delta x_{\text{goal}}, \Delta y_{\text{goal}})\) denotes the desired horizontal displacement of the root.
The orientation component \(q_{\text{goal}}\) specifies a desired yaw orientation represented as a \emph{relative} quaternion; importantly, this yaw is defined in the world frame and aligned with the box orientation, rather than with the reference motion.
The policy does not observe reference trajectories, future reference windows, or explicit phase variables; all task intent is conveyed solely through this goal specification.

\paragraph{Goal definition relative to the box.}
In box-based parkour tasks, the goal is defined relative to the box through a canonical, task-dependent location.
For the walk--jump and walk--climb tasks, the goal corresponds to the \emph{center of the top surface of the box}, expressed in the character-centric frame.
For the climb--down task, the goal corresponds to a target root location on the ground positioned at a fixed distance in front of the box (0.9\,m along the forward direction from the box center).
The policy is not explicitly provided with box geometry such as height or dimensions; instead, the box influences behavior implicitly through physical interaction and feasibility constraints.

\section{Reward Function}
\label{sec:sup_reward}

\paragraph{Tracking Rewards}
During imitation, we employ dense tracking rewards that encourage the humanoid to remain consistent with reference motion statistics without explicitly conditioning the policy on the reference.
Tracking rewards are implemented using exponentially weighted penalties of the form
\begin{equation}
r_{\text{track}} = \exp\!\left(-\kappa \frac{\|e\|^2}{\sigma_i^2}\right),
\end{equation}
where $e$ denotes a task-specific tracking error, $\sigma_i$ is a normalization scale, and $\kappa$ controls sensitivity.

We include tracking terms for base position and orientation, base linear and angular velocity, joint positions, and key body positions and orientations.
In addition, we explicitly encourage alignment between the robot’s projected gravity direction and the reference, which improves balance and orientation stability during dynamic motions.

\paragraph{Task Rewards}
For goal-conditioned generalization tasks, reference-dependent terms are removed and replaced with task-level objectives.
We include penalties on target base position and orientation errors relative to the goal, as well as a constant success reward that is activated when the robot reaches the target region.
These terms provide minimal task supervision, forcing the policy to reuse behaviors acquired during imitation.

\paragraph{Regularization}
To promote smooth and physically feasible behavior, we apply penalties on action smoothness, joint accelerations, applied torques, and violations of joint position and torque limits.
We further penalize large horizontal contact forces at the feet, foot slippage, excessive foot jerk, and undesirable ankle configurations.
These terms are critical for stabilizing training in contact-rich scenarios such as jumping and climbing.

\paragraph{Survival Reward}
To discourage early termination and promote sustained execution, we include a constant survival reward that is applied at every timestep.

All reward terms are summarized in Table~\ref{tab:reward_terms}, along with their weights and normalization constants.

\begin{table*}[t]
\centering
\small
\caption{Reward terms and hyperparameters used in training.
For exponential tracking terms, we use the form $\exp(-\kappa \|e\|^2/\sigma_i^2)$, where $\kappa$ is a global sensitivity parameter shared across all tracking terms and $\sigma_i$ is a per-term normalization scale.
We use $r_{IB}$ and $\Phi_{IB}$ to denote the base position and orientation in the world frame, $v_{IB}$ and $\omega_{IB}$ for base linear and angular velocities, and $q$ for joint positions; starred quantities $(\cdot)^\ast$ denote reference values.
The operator $\ominus$ denotes the $\mathrm{SO}(3)$ difference implemented via the rotation-log map.
For the joint position term, $n_j$ denotes the number of actuated joints.
For contact-related terms, let $F^{w}_{t,h,b}\in\mathbbm{R}^3$ denote the net contact force in the world frame acting on body $b$ at history index $h$, and let $\mathcal{B}$ denote the set of foot bodies.
Let $v^{w}_{\text{foot}}(b)\in\mathbbm{R}^2$ denote the planar velocity of foot $b$.
In the inequality ankle position limit penalty, $A$ and $b$ define a convex polytope constraint on ankle joint configurations.
In the flat ankle penalty, $\tilde{g}^{(i)}_z$ denotes the $z$-component of the gravity vector expressed in the aligned local frame of ankle $i$.
For the foot clearance reward, $h_b$ denotes the vertical height of foot $b$, $h^\ast$ is the target clearance height, and $\alpha$ is a scalar controlling velocity gating inside the $\tanh(\cdot)$ term.
Finally, $\Delta t$ denotes the simulation timestep used to compute foot jerk.}
\label{tab:reward_terms}
\begin{tabular}{l l c c}
\toprule
\textbf{Term Name} & \textbf{Definition (per-env scalar)} & \textbf{Weight} & $\boldsymbol{\sigma_i}$ \\
\midrule
\multicolumn{4}{c}{\textbf{Tracking Rewards}} \\
\midrule

Base position tracking
& $\exp\!\left(-\kappa \left\| r_{IB} - r^\ast_{IB} \right\|^2 / \sigma_1^2\right)$
& $1$ & $0.4$ \\

Base orientation
& $\exp\!\left(-\kappa \left\| \Phi_{IB} \ominus \Phi^\ast_{IB} \right\|^2 / \sigma_2^2\right)$
& $1$ & $0.5$ \\

Base angular velocity
& $\exp\!\left(-\kappa \left\| \omega_{IB} - \omega^\ast_{IB} \right\|^2 / \sigma_3^2\right)$
& $1$ & $1.5$ \\

Base linear velocity
& $\exp\!\left(-\kappa \left\| v_{IB} - v^\ast_{IB} \right\|^2 / \sigma_4^2\right)$
& $1$ & $0.6$ \\

Joint position
& $\exp\!\left(-\kappa \left\| q - q^\ast \right\|^2 / \sigma_5^2\right)$
& $1$ & $0.3\cdot \sqrt{n_j}$ \\

Base height tracking penalty
& $|z_{\text{base}} - z_{\text{ref}}|$
& $-10.0$ & -- \\

\midrule
\multicolumn{4}{c}{\textbf{Goal-Conditioned Task Rewards}} \\
\midrule

Target base position penalty
& $\left\|p^{xy}_{\text{base}} - p^{xy}_{\text{goal}}\right\|$
& $-5.0$ & -- \\

Target base orientation penalty
& $\left\|\Phi_{\text{base}} \ominus \Phi_{\text{goal}}\right\|$
& $-1.0$ & -- \\

Target reach reward
& $\mathbbm{1}[\text{reach}]$
& $10.0$ & -- \\

\midrule
\multicolumn{4}{c}{\textbf{Regularization and Contact Terms}} \\
\midrule

Large horizontal foot force (indicator)
& $\mathbbm{1}\!\left[\;\overline{f}^{xy}_{\max} > 10\;\right],\;
\overline{f}^{xy}_{\max}=\frac{1}{|\mathcal{B}|}\sum_{b\in\mathcal{B}}
\max_{h}\left\| (F^{w}_{t,h,b})_{xy}\right\|_2$
& $-10.0$ & -- \\

Action smoothness penalty
& $\left\|a_t-a_{t-1}\right\|_2$
& $-1.0$ & -- \\

Applied torque penalty
& $\left\|\tau_{\text{applied}}\right\|_2$
& $-5\times10^{-4}$ & -- \\

Joint position limit penalty
& $\sum_i \bigl[\max(0, q_i^{\min}-q_i) + \max(0, q_i-q_i^{\max})\bigr]$
& $-5.0$ & -- \\

Applied torque limit penalty
& $\sum_i \left|\tau_{\text{applied},i}-\tau_{\text{computed},i}\right|$
& $-0.1$ & -- \\

Inequality ankle position limit penalty
& $\mathrm{clamp}\!\left(\sum_k \max\!\left(0,\; (q_{\text{ankle}} A^\top)_k - b_k\right),\,10\right)$
& $-2.0$ & -- \\

Foot slip penalty
& $\mathrm{clamp}\!\left(\sum_{b\in\mathcal{B}} \left\|v^{w}_{\text{foot}}(b)\right\|_2\;\mathbbm{1}\!\left[\max_{h}\|F^{w}_{t,h,b}\|_2>1\right],\,10\right)$
& $-2.0$ & -- \\

Foot jerk penalty
& $\mathrm{clamp}\!\left(\left\|\frac{a^{w}_{\text{foot}}(t)-a^{w}_{\text{foot}}(t-1)}{\Delta t}\right\|_{F},\,10\right)$
& $-5\times10^{-4}$ & -- \\

Flat ankle penalty
& $\left(\tilde{g}^{(1)}_z + 1\right)^2 + \left(\tilde{g}^{(2)}_z + 1\right)^2$
& $-20.0$ & -- \\

Foot clearance reward
& $\exp\!\left(-\frac{1}{\sigma_6}\sum_{b\in\mathcal{B}}
\bigl(h_b-h^\ast\bigr)^2\;\tanh\!\left(\alpha \|v^{w}_{\text{foot}}(b)\|_2\right)\right)$
& $2.0$ & $0.05$ \\

\midrule
\multicolumn{4}{c}{\textbf{Survival Reward}} \\
\midrule

Survival bias
& $1$
& $30.0$ & -- \\

\bottomrule
\end{tabular}
\end{table*}

\section{Curriculum}
\label{sec:sup_curriculum}

We employ a wrench-based curriculum that stabilizes early-stage learning for dynamic contact-rich behaviors and also task mixing curriculum to gradually transfer emphasis from reference-guided imitation to goal-driven generalization. All curriculum-related parameters are summarized in Table~\ref{tab:sup_curriculum}.

\subsection{Assistive Wrench Curriculum}
A virtual assistive wrench is applied at the robot base during early training to reduce catastrophic failures.
Let $(p,v,\Phi,\omega)$ denote the base position, linear velocity, orientation, and angular velocity, and $(\hat{p},\hat{v},\hat{\Phi},\hat{\omega})$ their reference counterparts.
We compute a nominal spatial wrench using PD feedback on base tracking errors combined with feedforward terms.
The applied wrench is scaled by a difficulty-dependent factor $\beta(\lambda)$, which decreases monotonically as training progresses.

\subsection{Task Mixing Schedule}
The same scalar difficulty variable $\lambda \in [0,1]$ also controls the probability of sampling the imitation task versus the generalization task.
Training begins with imitation-dominated sampling and gradually shifts toward more frequent generalization-task sampling as the policy becomes more stable.

\begin{table*}[t]
\centering
\small
\caption{Curriculum hyperparameters.}
\label{tab:sup_curriculum}
\begin{tabular}{l l}
\toprule
\textbf{Component} & \textbf{Specification} \\
\midrule

Assistive wrench scaling
& $\beta(\lambda) = (1-\lambda)\,\beta_{\max}$ \\

Maximum assistive scale $\beta_{\max}$
& $0.75$ \\

Virtual force PD gains $(k_p^v, k_d^v)$
& $(0,15)$ \\

Virtual torque PD gains $(k_p^\omega, k_d^\omega)$
& $(200,1)$ \\

\midrule
Imitation sampling probability
& $p_{\text{imi}}(\lambda) = (1-\lambda)p_0 + \lambda p_{\text{target}}$ \\

Initial imitation probability $p_0$
& 1.0 \\

Final imitation probability $p_{\text{target}}$
& $0.5$ \\

\bottomrule
\end{tabular}
\end{table*}

\section{Network Architecture and Training Details}
\label{sec:sup_mdp_ppo}

This section summarizes the Markov Decision Process (MDP) configuration, network architecture, and optimization hyperparameters used in all experiments.
Unless otherwise noted, the same settings are shared across tasks and behaviors.
We report all relevant hyperparameters explicitly to facilitate reproducibility in Table~\ref{tab:sup_mdp_ppo}.

\begin{table*}[t]
\centering
\small
\caption{MDP configuration, network architecture, and PPO hyperparameters.}
\label{tab:sup_mdp_ppo}
\begin{tabular}{l l}
\toprule
\textbf{Hyperparameter} & \textbf{Value} \\
\midrule
\multicolumn{2}{c}{\textbf{MDP and Simulation Setup}} \\
\midrule
Episode length ($\bar{L}_{\text{episode}}$) & $10.0$ s \\
Simulation time-step ($dt$) & $0.004$ s \\
Control decimation & $5$ \\
Control frequency & $50$ Hz \\
\midrule
\multicolumn{2}{c}{\textbf{Policy and Value Networks}} \\
\midrule
Actor network & MLP$(1024,\;512,\;256)$ with ELU activations \\
Critic network & MLP$(1024,\;512,\;256)$ with ELU activations \\
Actor observations & Proprioception + goal \\
Critic observations & Actor obs + privileged information \\
\midrule
\multicolumn{2}{c}{\textbf{PPO Hyperparameters}} \\
\midrule
Learning rate (start of training) & $1\times10^{-4}$ \\
Discount factor ($\gamma$) & $0.99$ \\
GAE discount factor ($\lambda_{\text{GAE}}$) & $0.95$ \\
Desired KL-divergence & $0.01$ \\
PPO clip range & $0.2$ \\
Entropy coefficient & $0.001$ \\
Value function loss coefficient & $0.5$ \\
Number of epochs per update & $5$ \\
Number of environments & $4096$ \\
Batch size & $245{,}760$ $(4096 \times 24)$ \\
Mini-batch size & $61{,}440$ $(4096 \times 6)$ \\
\bottomrule
\end{tabular}
\end{table*}

\section{State Initialization and Domain Randomization}
\label{sec:sup_dr}

We adopt Reference State Initialization(RSI)~\cite{peng2018deepmimic}. To encourage robustness and prevent overfitting to the exact reference trajectory, we apply randomized perturbations to the initialized base pose. These perturbations are applied across the entire reference trajectory.

For the walk--climb and walk--jump tasks, we apply larger perturbations to distances and orientations relative to the box. For the climb--down task, smaller perturbations are used to maintain feasibility near the top of the box. All perturbation ranges are summarized in Table~\ref{tab:sup_init_offsets}.

\begin{table*}[t]
\centering
\small
\caption{Reference-based initialization perturbations applied to the base pose. All offsets are sampled uniformly within the specified ranges and applied relative to the reference state.}
\label{tab:sup_init_offsets}
\begin{tabular}{l c c c c}
\toprule
\textbf{Task} 
& $\Delta x$ (m) 
& $\Delta y$ (m) 
& Yaw (rad) 
& Roll / Pitch (rad) \\
\midrule
Walk--jump 
& $\pm 0.4$ 
& $\pm 0.4$ 
& $\pm 0.8$ 
& $\pm 0.15$ \\

Walk--climb 
& $\pm 0.4$ 
& $\pm 0.4$ 
& $\pm 0.8$ 
& $\pm 0.15$ \\

Climb--down 
& $\pm 0.2$ 
& $\pm 0.2$ 
& $\pm 0.6$ 
& $\pm 0.15$ \\
\bottomrule
\end{tabular}
\end{table*}

To improve robustness and facilitate transfer, we apply domain randomization during training, including perturbations to dynamics, contacts, and observations. All domain randomization ranges are summarized in Table~\ref{tab:sup_domain_rand}.

\begin{table*}[t]
\centering
\small
\caption{Domain randomization terms used during training.}
\label{tab:sup_domain_rand}
\begin{tabular}{l l}
\toprule
\textbf{Term} & \textbf{Value} \\
\midrule
Static friction
& $\mathcal{U}(0.8,\;2.5)$ \\

Dynamic friction
& $\mathcal{U}(0.7,\;2.5)$ \\

Restitution
& $\mathcal{U}(0.0,\;0.2)$ \\

Torso mass
& $\text{default} + \mathcal{U}(-2.5,\;4.0)\ \text{kg}$ \\

Pelvis mass
& $\text{default} + \mathcal{U}(-1.0,\;1.0)\ \text{kg}$ \\

External disturbance (impulsive push at base)
& Interval $=\mathcal{U}(0.0\,\text{s},\;4.0\,\text{s})$, \quad
$v_{xy}=0.4\ \text{m\,s}^{-1}$ \\

\bottomrule
\end{tabular}
\end{table*}

\section{Reference data}
We obtain the reference motions from a single demonstration for each skill, recorded using a standard iPhone. From the captured video, we reconstruct both the static scene geometry and an aligned 3D human motion sequence using CRISP~\cite{wang2025crisp}, which provides a consistent human–scene reconstruction from monocular input. The reconstructed human motion is then retargeted to the humanoid robot using GMR~\cite{araujo2025retargeting}, mapping the human kinematics to the robot. The resulting retargeted motion is used to define tracking rewards and goal conditions during training, and is not provided as input to the policy at execution time. For the walk–climb and walk–jump skills, we additionally augment the training data by mirroring the demonstration.


\end{document}